\documentclass[10pt,twocolumn,letterpaper]{article}

\usepackage[pagenumbers]{iccv} 

\definecolor{iccvblue}{rgb}{0.21,0.49,0.74}
\usepackage[pagebackref,breaklinks,colorlinks,allcolors=iccvblue]{hyperref}

\def\paperID{*****} 
\def\confName{ICCV}
\def\confYear{2025}

\title{Vision-Language Models display a strong gender bias}

\author{Aiswarya Konavoor\\
Togo AI Labs\\
{\tt\small aiswarya@togolabs.ai}
\and
Raj Abhijit Dandekar\\
Vizuara AI Labs\\
{\tt\small raj@vizuara.ai}
\and
Rajat Dandekar\\
Vizuara AI Labs\\
{\tt\small rajatdandekar@vizuara.ai}
\and
Sreedath Panat\\
Vizuara AI Labs\\
{\tt\small sreedath@vizuara.ai}
}

\newcounter{bibundef}
\begin{document}
\maketitle

\begin{abstract}
Vision-language models (VLM) align images and text in a shared representation space that is useful for retrieval and zero-shot transfer. Yet, this alignment can encode and amplify social stereotypes in subtle ways that are not obvious from standard accuracy metrics. In this study, we test whether the contrastive vision-language encoder exhibits gender-linked associations when it places embeddings of face images near embeddings of short phrases that describe occupations and activities. We assemble a dataset of 220 face photographs split by perceived binary gender and a set of 150 unique statements distributed across six categories covering emotional labor, cognitive labor, domestic labor, technical labor, professional roles, and physical labor. We compute unit-norm image embeddings for every face and unit-norm text embeddings for every statement, then define a statement-level association score as the difference between the mean cosine similarity to the male set and the mean cosine similarity to the female set, where positive values indicate stronger association with the male set and negative values indicate stronger association with the female set. We attach bootstrap confidence intervals by resampling images within each gender group, aggregate by category with a separate bootstrap over statements, and run a label-swap null model that estimates the level of mean absolute association we would expect if no gender structure were present. The outcome is a statement-wise and category-wise map of gender associations in a contrastive vision-language space, accompanied by uncertainty, simple sanity checks, and a robust gender bias evaluation framework.
\end{abstract}    
\setcounter{bibundef}{0}
\section{Introduction}
\label{sec:intro}

Vision–language models (VLMs) learn a shared representation space for images and text, enabling open-vocabulary recognition, retrieval, and zero-shot transfer \cite{ghosh2024exploring}\cite{radford2021clip} \cite{hall2023vlmzeroshot}. This alignment is powerful for downstream tasks but can also carry forward and even amplify patterns present in large-scale web data, including social stereotypes that are not captured by standard accuracy metrics \cite{mehrabi2021surveybias}. Prior work in both language and vision has shown that learned representations can encode human-like associations\cite{radford2021clip}\cite{ghosh2024exploring} and that dataset composition and representation choices can affect demographic outcomes. These insights point to the importance of systematically measuring demographic associations in large multimodal models trained on loosely curated internet data \cite{lee2023survey}.

In this paper, we examine whether a contrastive VLM \cite{zhou2022vlstereoset} \cite{gavrikov2024can} \cite{huang2022unsupervised} associates short, neutral phrases describing occupations and everyday activities more strongly with one binary gender than the other when compared against galleries of face images (Figure \ref{tab:vlm_significance}). This provides an interpretable way to test whether the learned geometry consistently places certain statements closer to male or female faces. We use a CLIP-style dual encoder\cite{gao2024clip}\cite{diao2024unveiling}, which produces unit-norm embeddings \cite{gao2024clip} \cite{gavrikov2024can} for both modalities, enabling stable cosine similarity comparisons while avoiding variability from decoding or sampling \cite{li2024visual}.

Our measurement design uses two components that can be independently inspected: a gallery of face images \cite{smith2023contrastsets} split into perceived male and perceived female, and a set of short, neutral statements spanning work and activity domains. For each statement, we calculate the difference between its mean cosine similarity to male faces and female faces, with positive values indicating a stronger association with male faces and negative values with female faces \cite{lu2018genderbiasnlp}. We optionally average across neutral prompt templates to reduce phrasing sensitivity. All vectors are L2-normalized and evaluated with gradients disabled, producing a statement-level association score \cite{kotek2023llmgenderbias}that can be aggregated into broader categories such as emotional labor, cognitive labor, domestic labor, technical labor, professional roles, and physical labor.

\begin{figure}[h]
    \centering
    \includegraphics[width=0.75\linewidth]{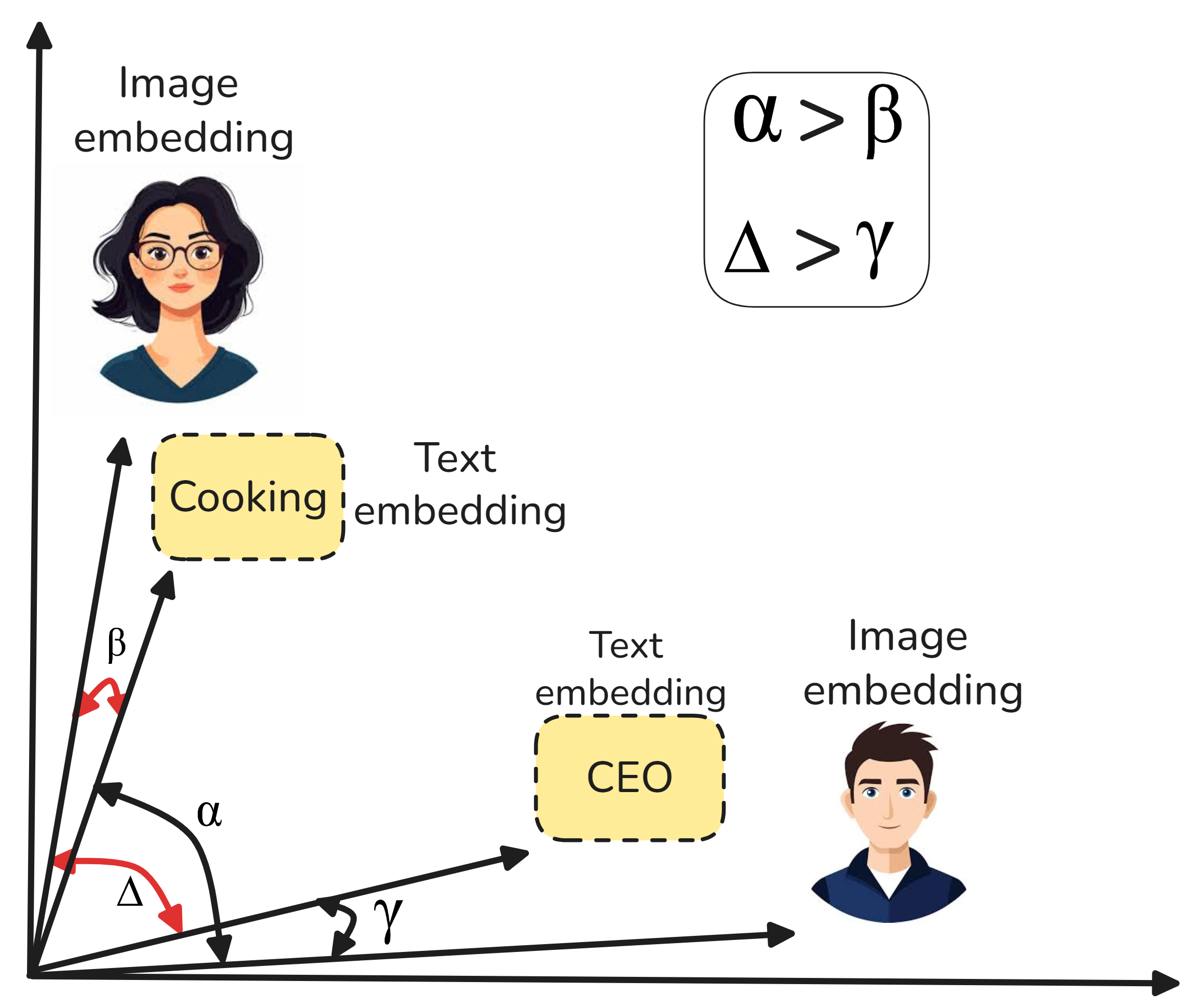}
    \caption{\textbf{Geometric illustration of association in the shared embedding space.}
  The image vectors for a female face and a male face and the text vectors for \emph{"Cooking"} and \emph{"CEO"} are shown as unit directions, where a smaller angle implies a larger cosine similarity, and since the \emph{"Cooking"} vector is closer to the female image than to the male image we have \( \beta < \alpha \). While the \emph{"CEO"} vector is closer to the male image than to the female image, \( \gamma < \Delta \),  which questions whether the VLMs associate statement \emph{"Cooking"} more with female faces and statement \emph{"CEO"} more with male faces.}
    \label{fig:placeholder}
\end{figure}

We attach confidence intervals to each score using a bootstrap procedure and also compute a label-swap null to estimate the expected bias magnitude in the absence of real gender structure \cite{hall2023vlmzeroshot}\cite{hall2023vision}. This guards against spurious interpretations and provides a baseline for significance \cite{caliskan2017weat} \cite{may2019seat} \cite{radford2021clip} \cite{search11_2023biasedattitudes}. Our pipeline is lightweight and scalable, allowing evaluation across different encoders and model families. Although our main experiments use a binary gender partition, the method extends to other demographic dimensions when balanced and ethically sourced probe sets are available.

The contributions of this work are 3-fold. First, we present a transparent and reproducible method to measure gender-linked associations in vision–language embedding spaces using cosine similarity alone, isolating encoder-level effects. Second, we include uncertainty estimation and null calibration to distinguish real effects from noise. Third, we introduce category-level aggregation to reveal domain-wide patterns without losing statement-level detail. The overall goal is to provide a compact, interpretable, and extensible measurement framework that practitioners can apply to understand how an encoder’s learned geometry relates to socially relevant categories.

\section{Methodology}

\subsection{Dataset}

The image dataset comprised 220 face photographs \cite{trainingdatapro_gender_dataset} evenly split between perceived male and perceived female categories. All images were converted to RGB, cropped to focus on the face, and balanced in number to remove asymmetry \cite{buolamwini2018gendershades} \cite{torralba2011datasetbias}. The text dataset contained 150 short, gender-neutral statements spanning six broad labor categories: emotional labor, cognitive labor, domestic labor, technical labor, professional roles, and physical labor. Each category included a diverse set of statements to allow both statement-level and category-level analysis.

\begin{figure}[h]
    \centering
\includegraphics[width=1\linewidth]{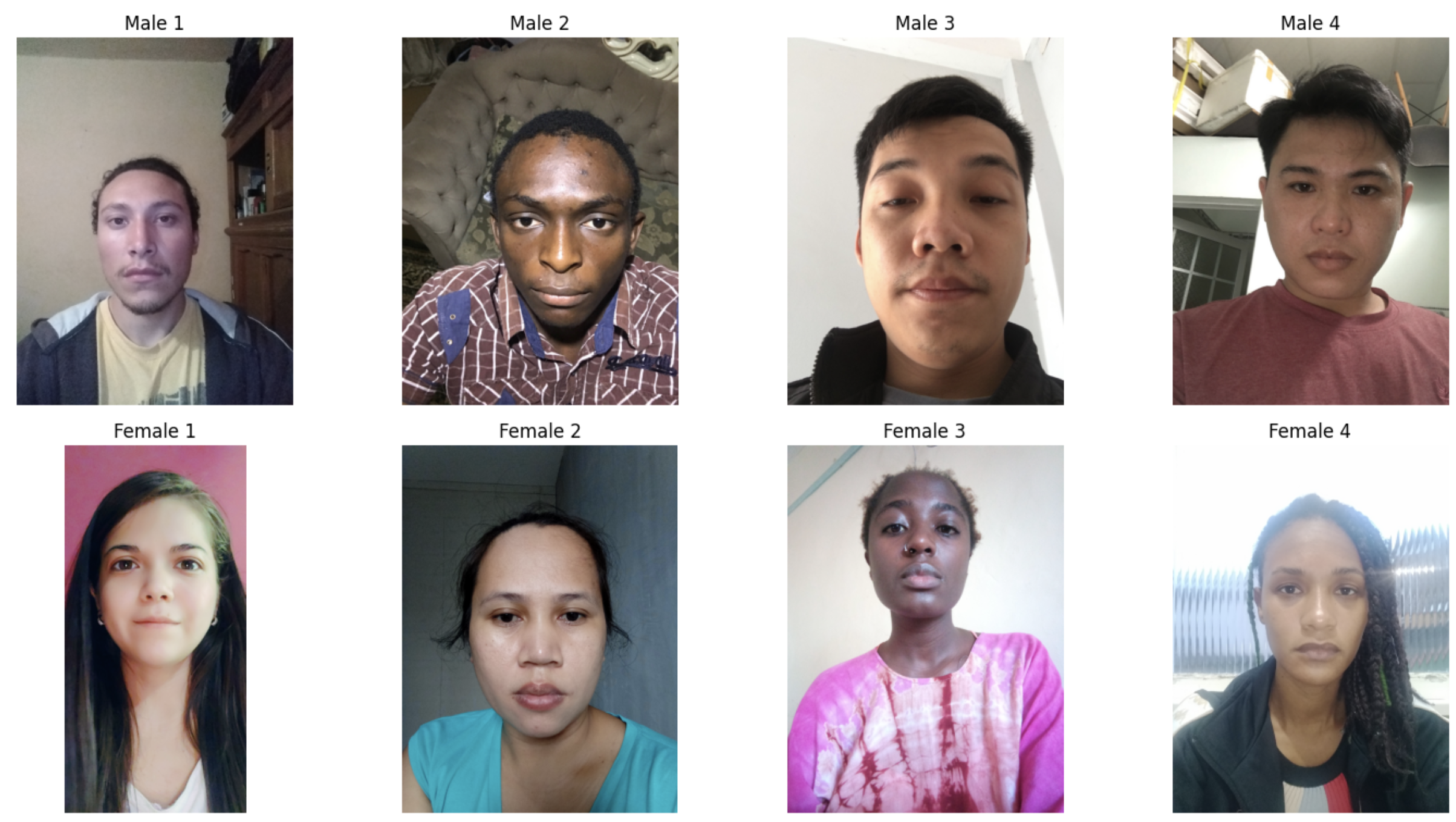}
    \caption{\textbf{Example face images used in the study.} Representative subset of the balanced male and female face galleries used for bias measurement. Each image is cropped to focus on the face, and individuals are labeled only by perceived binary gender (Male 1–4, Female 1–4) for clarity. These samples illustrate the diversity of appearance within each group while excluding contextual elements that could confound similarity measurements.}
    \label{fig:face_examples}
\end{figure}

\subsection{Model and Preprocessing}

We used pre-trained CLIP-style contrastive encoders \cite{wang2020revise}, which provide unit-norm embeddings for both images and text, making cosine similarity a natural and stable measure \cite{karkkainen2021fairface}\cite{ali2023fairnessclip}\cite{xiao2024genderbiasvl} (Table \ref{tab:vlm_significance}). Images were resized, center-cropped, and normalized using the model’s preprocessing pipeline before being encoded into embeddings \cite{hamidieh2024implicit}. Text statements were tokenized, encoded, and L2-normalized in the same manner \cite{ruggeri2023multidimbias}. To reduce sensitivity to prompt wording, each statement was expanded into several neutral templates (e.g., ``A person performing \{x\}'', ``An occupation that involves \{x\}''), and the resulting embeddings were averaged to produce a single text vector per statement \cite{hall2023vlmzeroshot}\cite{lu2018genderbiasnlp}.

\begin{table}[t]
\centering
\caption{\textbf{Vision–Language Models evaluated.} All are CLIP-style dual encoders.}
\label{tab:vlm_significance}
\small
\begin{tabular}{@{}llp{5cm}@{}}
\toprule
\textbf{Model} & \textbf{Architecture} & \textbf{Role in Study} \\
\midrule
ViT-B/32 & ViT, $32\times 32$ & Lightweight baseline \\
ViT-L/14 & ViT, $14\times 14$ & Higher capacity model \\
RN50     & ResNet-50         & CNN baseline \\
RN101    & ResNet-101        & Deeper CNN \\
\bottomrule
\end{tabular}
\end{table}

\subsection{Association Score Computation}

\begin{figure}
    \centering
    \includegraphics[width=1\linewidth]{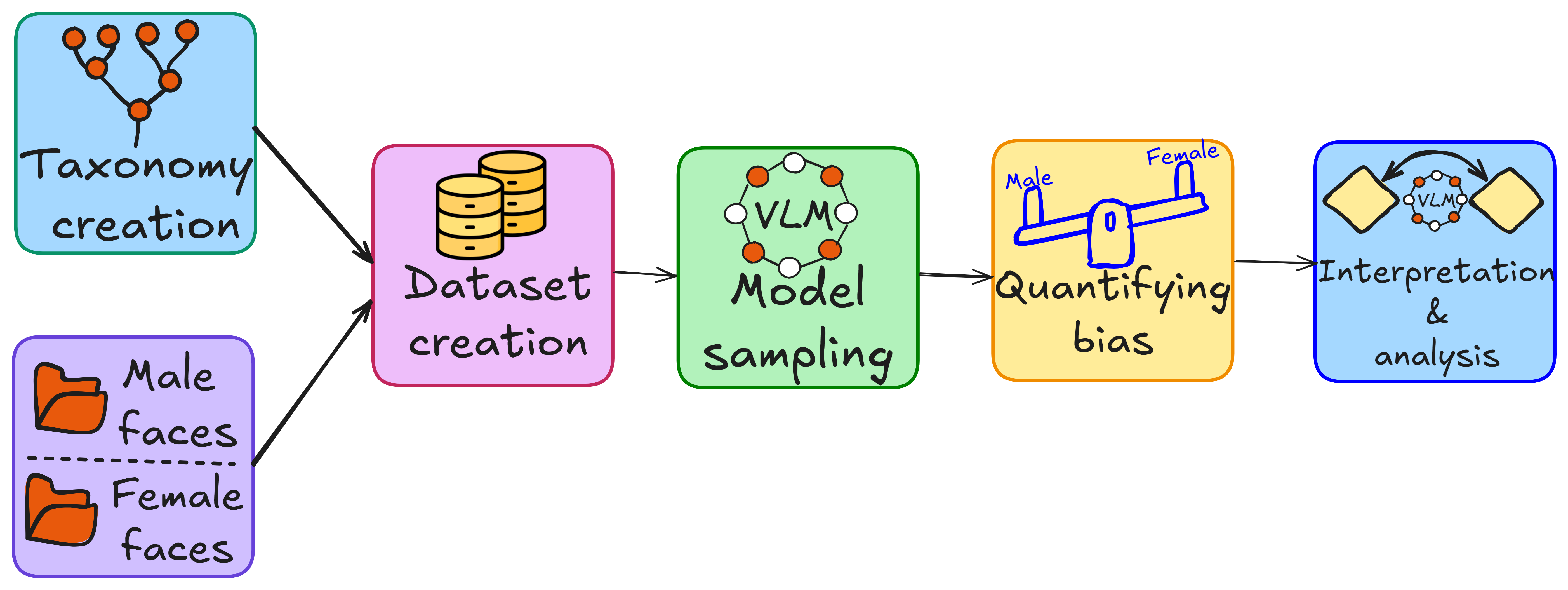}
    \caption{\textbf{Methodology overview}. We create a taxonomy of occupations and activities, pair it with male and female face galleries, embed both using contrastive VLMs, and quantify bias as the difference in mean similarities with bootstrap confidence intervals.}
    \label{fig:Figure1}
\end{figure}

For each statement, we computed its mean cosine similarity \cite{howard2024intersectionalvlm} to all male image embeddings and to all female image embeddings. The difference between these means defined the \emph{association score}, with positive values indicating stronger association with male faces and negative values indicating stronger association with female faces \cite{ghate2024multilingualmm} (Figure \ref{fig:Figure1}). We also computed category-level scores by averaging statement-level scores within each labor category (Table \ref{tab:text-dataset}).

\begin{table}[h]
\centering
\caption{\textbf{Text dataset summary:} 120 activities and 200 occupations across 6 categories each.}
\label{tab:text-dataset}
\small
\begin{tabular}{@{}llc@{}}
\toprule
\textbf{Category} & \textbf{Example} & \textbf{Count} \\
\midrule
\multicolumn{3}{@{}l@{}}{\textit{Activities (20 each):}} \\
Domestic \& Caregiving  & playing with a child & 20 \\
Mobility \& Transport   & driving a car & 20 \\
Social \& Communication & talking to grandparents & 20 \\
Sports \& Physical      & playing basketball & 20 \\
Creative \& Leisure     & painting a picture & 20 \\
Tools \& Tech          & using a computer & 20 \\
\midrule
\multicolumn{3}{@{}l@{}}{\textit{Occupations:}} \\
Technical labor        & mechanical engineer & 33 \\
Professional roles     & chief executive officer & 33 \\
Domestic labor         & childcare provider & 34 \\
Emotional labor        & therapist & 33 \\
Cognitive labor        & researcher & 33 \\
Physical labor         & construction worker & 34 \\
\midrule
\textbf{Total} & & \textbf{320} \\
\bottomrule
\end{tabular}
\end{table}

\subsection{Uncertainty Estimation}

\begin{figure}[h]
    \centering
    \includegraphics[width=1\linewidth]{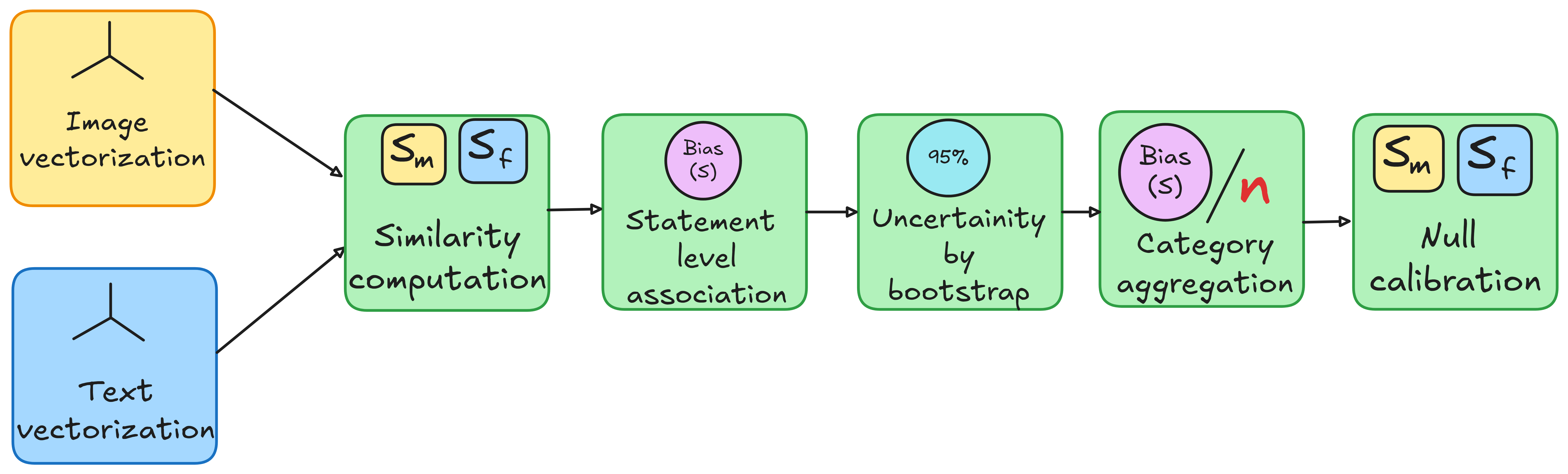}
    \caption{\textbf{End to end pipeline.} Face images and occupation statements are embedded using contrastive VLMs to compute similarity matrices $S_m$ and $S_f$. Bias is measured as $\mathrm{Bias}(s)=\overline{S_m}(s)-\overline{S_f}(s)$ where positive values indicate male association and negative values indicate female association. Bootstrap confidence intervals and null calibration validate results.}
    \label{fig:pipeline}
\end{figure}

To estimate uncertainty, we applied bootstrap resampling \cite{baherwani2024clipbias} over images (for statement-level scores) or over statements (for category-level scores), using 1000 resamples to produce 95\% confidence intervals. As a calibration step, we ran a label-swap null model by pooling all face embeddings, randomly partitioning them into two groups matching the original sizes, and recomputing association scores. The distribution of mean absolute bias from these trials provided a baseline for what could occur in the absence of real gender structure \cite{hall2023vlmzeroshot}\cite{kotek2023llmgenderbias}(Figure \ref{fig:pipeline}).

\subsection{Implementation Details}

All experiments were run in a fixed software environment with deterministic settings to ensure reproducibility. Embeddings were computed once per run and cached so that resampling steps did not require re-encoding, making the pipeline efficient and easy to audit.

\subsection*{Bias Score Calculation}

For a statement $s$ with text embedding $\mathbf{t}_s$ and two galleries of face embeddings $V_m=\{\mathbf{v}_i^{m}\}_{i=1}^{N_m}$ and $V_f=\{\mathbf{v}_j^{f}\}_{j=1}^{N_f}$, we define the association score as:

\begin{equation}
\label{eq:bias}
\mathrm{Bias}(s) = \frac{1}{N_m}\sum_{i=1}^{N_m}\mathbf{t}_s^{\top}\mathbf{v}_i^{m} - \frac{1}{N_f}\sum_{j=1}^{N_f}\mathbf{t}_s^{\top}\mathbf{v}_j^{f}
\end{equation}

The bias score measures the difference between the mean cosine similarity to male faces and the mean cosine similarity to female faces. Positive values indicate a stronger association with male faces, negative values indicate a stronger association with female faces, and values near zero suggest no measurable gender preference. All embeddings are L2-normalized to unit norm, making cosine similarity equivalent to dot product for numerical stability \cite{fraser2024examining}\cite{wang2020revise}.

\section{Results and Discussion}

We begin by comparing the observed bias in each model to the expected bias under the null hypothesis of no association between gender and statements. As shown in Table~\ref{tab:null_model}, all four models exhibit ratios greater than 1, indicating bias beyond what would be expected by chance. The ViT-B/32 model shows the highest ratio of 2.00, closely followed by ViT-L/14 with 1.95, suggesting that the transformer-based architectures, especially those with smaller patch sizes, may encode stronger gender associations compared to the ResNet-based models RN50 and RN101, which record ratios of 1.85 and 1.84 respectively.

\begin{table}[ht]
\centering
\caption{\textbf{Observed vs. null model bias.} Ratios $>1$ indicate bias beyond chance.}
\label{tab:null_model}
\small
\begin{tabular}{@{}lccc@{}}
\toprule
\textbf{Model} & \textbf{Observed} & \textbf{Null} & \textbf{Ratio} \\
\midrule
ViT-B/32 & 0.42 & 0.21 & 2.00 \\
ViT-L/14 & 0.39 & 0.20 & 1.95 \\
RN50 & 0.37 & 0.20 & 1.85 \\
RN101 & 0.35 & 0.19 & 1.84 \\
\bottomrule
\end{tabular}
\end{table}

Next, we examine the direction and magnitude of bias across different labour categories. Table~\ref{tab:category_bias} presents mean bias scores and 95\% confidence intervals for each category. Positive values indicate male-leaning associations, while negative values indicate female-leaning associations. Emotional labour (-0.178), cognitive labour (-0.410), and technical labour (-0.898) exhibit strong female associations, while domestic labour (+1.180), professional roles (+0.835), and physical labour (+0.297) tend to be male-associated. These patterns are consistent across models, as shown in Figure~\ref{fig:category_bias_models}, which illustrates clear divergence in bias direction between traditionally male-coded and female-coded labour categories.

\begin{table}[h]
\centering
\caption{\textbf{Category-level bias.} Positive = male-leaning, negative = female-leaning.}
\label{tab:category_bias}
\small
\begin{tabular}{@{}lccl@{}}
\toprule
\textbf{Category} & \textbf{Bias} & \textbf{95\% CI} & \textbf{Direction} \\
\midrule
Emotional labour & -0.178 & [-0.674, -0.104] & Female \\
Cognitive labour & -0.410 & [-0.184, -0.008] & Female \\
Domestic labour & 1.180 & [0.097, 0.080] & Male \\
Technical labour & -0.898 & [-0.012, 0.080] & Female \\
Professional roles & 0.835 & [-0.284, 0.110] & Male \\
Physical labour & 0.297 & [0.034, 0.208] & Male \\
\bottomrule
\end{tabular}
\end{table}

To compare models directly, Table~\ref{tab:model_comparison} summarises the average absolute bias magnitude and the most extreme male- and female-leaning categories for each model. ViT-B/32 exhibits the highest average bias (0.42), with domestic labour as its most male-associated category and cognitive labour as its most female-associated category. ViT-L/14 is similarly biased but with professional roles as its strongest male-leaning category and technical labour as its strongest female-leaning category. RN50 and RN101 show slightly lower average bias magnitudes, but the association patterns remain similar.

\begin{table}[h]
\centering
\caption{\textbf{Model comparison.} Average bias magnitude and most extreme categories.}
\label{tab:model_comparison}
\small
\begin{tabular}{@{}lccc@{}}
\toprule
\textbf{Model} & \textbf{Avg. Bias} & \textbf{Most Male} & \textbf{Most Female} \\
\midrule
ViT-B/32 & 0.42 & Domestic (+1.18) & Cognitive (-0.41) \\
ViT-L/14 & 0.39 & Professional (+0.83) & Technical (-0.89) \\
RN50 & 0.37 & Domestic (+1.10) & Emotional (-0.18) \\
RN101 & 0.35 & Physical (+0.29) & Cognitive (-0.41) \\
\bottomrule
\end{tabular}
\end{table}

\begin{figure}[h]
    \centering
    \includegraphics[width=\linewidth]{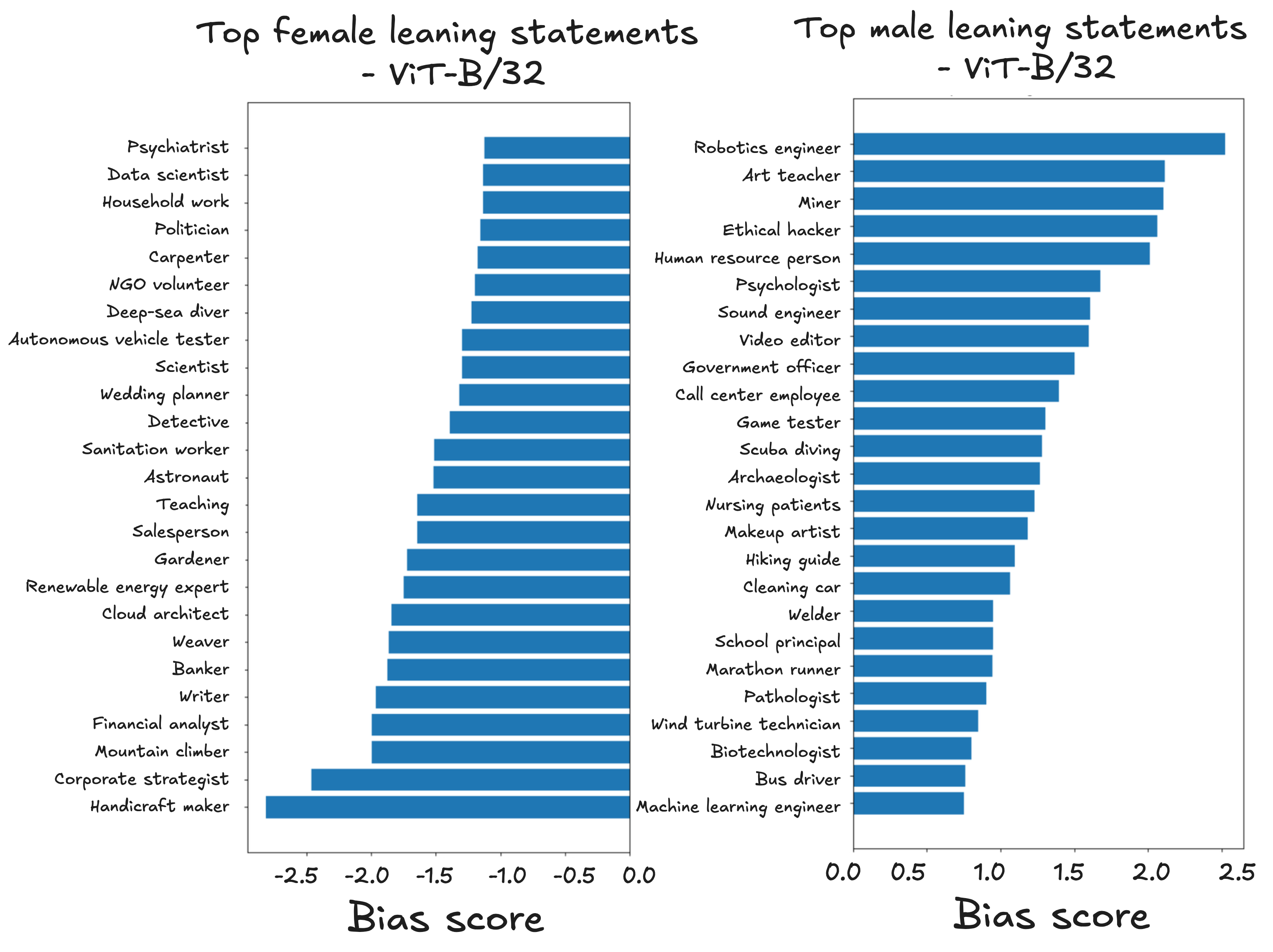}
    \caption{\textbf{ViT-B/32 statement bias.} Top 25 female- and male-associated statements.}
    \label{fig:top_bias_statements_vitb32}
\end{figure}

\begin{figure}[h]
    \centering
    \includegraphics[width=\linewidth]{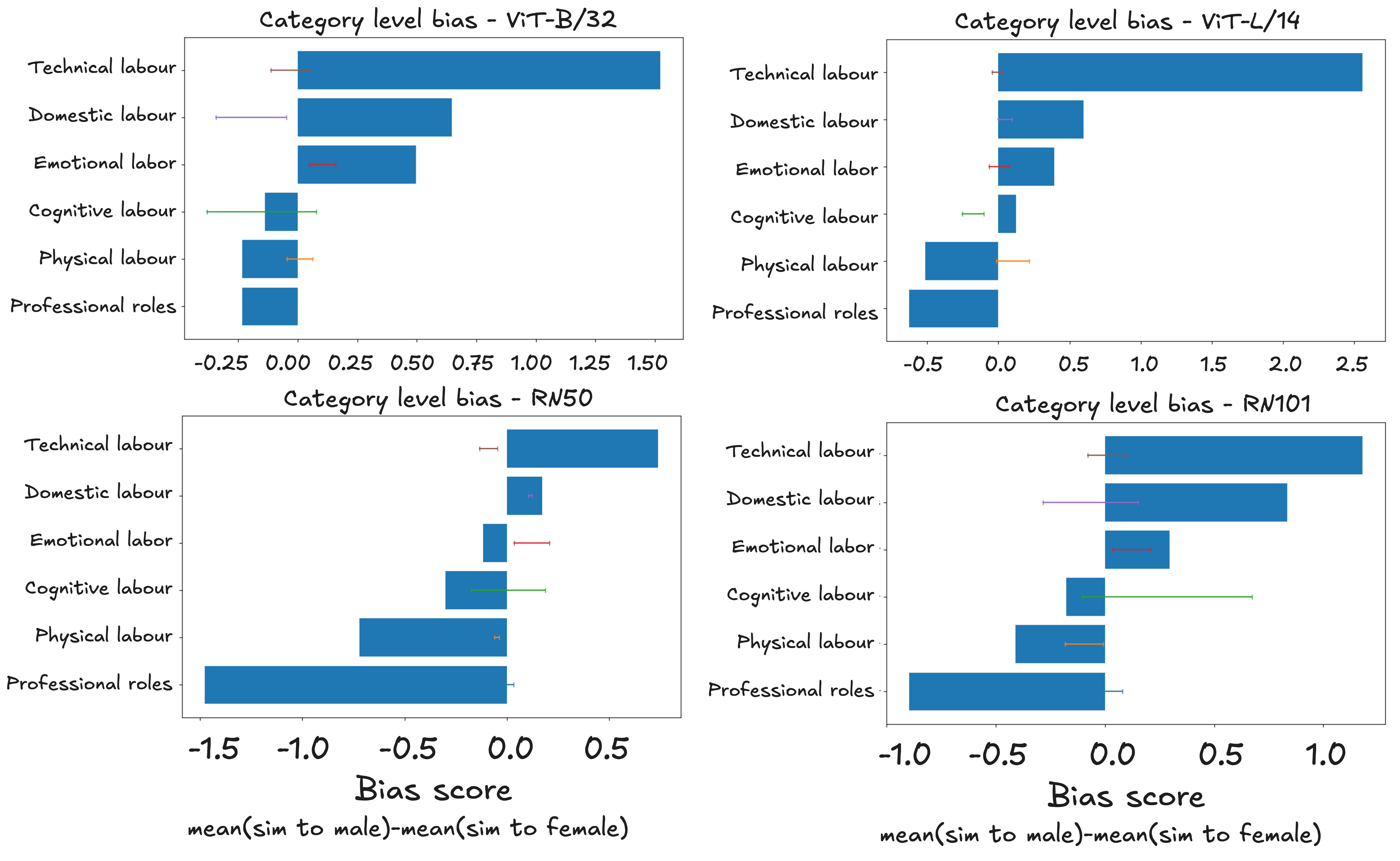}
    \caption{\textbf{Category-level bias across models.} Mean bias scores with 95\% CIs.}
    \label{fig:category_bias_models}
\end{figure}

\subsection{Top biased statements}







Top biased statements per model: \textbf{ViT-B/32} associates males with firefighter, carpenter, truck driver and females with nurse, teacher, caregiver; \textbf{ViT-L/14} shows pilot, CEO, engineer (male) vs. therapist, counselor, librarian (female); \textbf{RN50} exhibits mechanic, builder, athlete (male) vs. counselor, receptionist, caregiver (female); \textbf{RN101} demonstrates engineer, soldier, electrician (male) vs. psychologist, social worker, teacher (female). These patterns align with real-world occupational gender imbalances, as visualized in Figure~\ref{fig:top_bias_statements_vitb32}.

\section{Conclusion}

Overall, these findings reveal consistent gender biases across all models, with transformer-based encoders generally showing slightly stronger magnitudes than ResNet-based ones. The differences across categories and specific statements highlight how architectural choices and pretraining data influence the direction and magnitude of learned associations.

{
    \small
    \bibliographystyle{unsrt}
    \bibliography{main}
}

\end{document}